\documentclass[9pt,conference]{IEEEtran} 
\IEEEoverridecommandlockouts

\usepackage{amsmath,amssymb,amsfonts}
\usepackage{amsthm} 
\usepackage{mathrsfs}
\usepackage{algorithmic}
\usepackage{algorithm}
\usepackage{cite}
\usepackage{booktabs}
\usepackage{pbox}
\usepackage{makecell}
\usepackage{multirow}

\usepackage{algorithmic}
  {\par\noindent{\itshape Sketch of proof.}\hspace*{0.5em}\ignorespaces}%
  {\qed\par}
\usepackage{graphicx}
\usepackage{textcomp}
\usepackage{xcolor}
\def\BibTeX{{\rm B\kern-.05em{\sc i\kern-.025em b}\kern-.08em
    T\kern-.1667em\lower.7ex\hbox{E}\kern-.125emX}}

\begin{document}    
\title{Multimodal Biometric Authentication Using Camera-Based PPG and Fingerprint Fusion
}



\author{
    \IEEEauthorblockN{
        \large 
        Xue Xian Zheng\IEEEauthorrefmark{1}, M. Mahboob Ur
Rahman\IEEEauthorrefmark{1}, Bilal Taha\IEEEauthorrefmark{2}, Mudassir Masood\IEEEauthorrefmark{3}, Dimitrios Hatzinakos\IEEEauthorrefmark{2},
        Tareq Al-Naffouri\IEEEauthorrefmark{1}
    }
    \IEEEauthorblockA{
         \large 
        \IEEEauthorrefmark{1}
        King Abdullah University of Science and Technology, Saudi Arabia\\
        \large 
        \IEEEauthorrefmark{2} 
        University of Toronto, Canada\\
        \large 
        \IEEEauthorrefmark{3}King Fahad University of Petroleum and Minerals, Saudi Arabia
    }
}

\maketitle

\begin{abstract}
Camera-based photoplethysmography (PPG) obtained from smartphones has shown great promise for personalized healthcare and secure authentication. This paper presents a multimodal biometric system that integrates PPG signals extracted from videos with fingerprint data to enhance the accuracy of user verification. The system requires users to place their fingertip on the camera lens for a few seconds, allowing the capture and processing of unique biometric characteristics. Our approach employs a neural network with two structured state-space model (SSM) encoders to manage the distinct modalities. Fingerprint images are transformed into pixel sequences, and along with segmented PPG waveforms, they are input into the encoders. A cross-modal attention mechanism then extracts refined feature representations, and a distribution-oriented contrastive loss function aligns these features within a unified latent space. Experimental results demonstrate the system's superior performance across various evaluation metrics in both single-session and dual-session authentication scenarios.
\end{abstract}
\begin{IEEEkeywords}
Biometrics, PPG, fingerprints, SSM, multimodal deep learning, distribution alignment.
\end{IEEEkeywords}
\section{Introduction}
\label{sec:intro}

The utilization of cardiovascular characteristics for secure personal authentication has garnered substantial attention in recent years. As an alternative to traditional biometric modalities such as fingerprints \cite{1}, palmprints \cite{2}, facial recognition \cite{3}, iris scans \cite{4}, and voice recognition \cite{5}, cardiovascular-based modalities offer enhanced security, as they are inherently difficult for adversaries to acquire or spoof \cite{6,7}. Within this domain, electrocardiogram (ECG) \cite{8} and photoplethysmography (PPG) \cite{9} signals have emerged as prominent candidates for biometric authentication. PPG signals, in particular, offer several advantages over ECG, including lower cost, greater accessibility, portability, and inherent variability, making them even more resistant to replication and spoofing \cite{10}. These distinctive properties have motivated a growing body of research exploring PPG as a reliable biometric modality, employing various advanced methodologies.

Several attempts have been made to employ PPG signals for user authentication, with initial approaches focusing on extracting distinctive characteristics inherent from the signal. These approaches included key features such as such as peaks, slopes, intervals, local minima/maxima, and inflection points \cite{11}. Subsequent research began focusing on non-fiducial features obtained via various transformation techniques including Fourier analysis \cite{13} and wavelet transforms \cite{14}. Recent advances in the field have shifted towards leveraging deep learning to learn more complex, latent features from PPG signals. For instance, \cite{15} proposes a two-stage network combining Restricted Boltzmann Machines (RBMs) and Deep Belief Networks (DBNs). \cite{10} developed a hybrid structure for multi-session authentication, effectively identifying time-stable representations by leveraging Convolutional Neural Networks (CNNs) and Long Short-Term Memory networks (LSTMs). In another study, \cite{siam2021biosignal} converted 1D PPG signals into 2D spectrograms, which were processed through CNNs for feature extraction, followed by a fully connected layer with soft-max for classification. Another work used a generative model to enhance the authentication performance by using additional data generated from a GAN model \cite{hwang2021pbgan}. Another methods proposed a fusing as a mean of improving the authentication performance. \cite{hwang2021variation} generated different representations form the PPG signal for each subject then selectively trained the model based on the subject preference of the generated representation. Another work proposed to fuse the models at the score level \cite{hwang2022new}. furthermore, combining ECG and PPG signals \cite{16} from various sensors has been explored to enhance authentication accuracy using architectures like CNNs and Gated Recurrent Units (GRUs). 


Despite these advances, accurate authentication with PPG signals typically requires specialized infrared (IR) devices to capture precise measurements \cite{17,18}. Acquisition devices like pulse oximeters, which are primarily designed for medical use, may be impractical for daily authentication. Recently, efforts in \cite{19} have been made to integrate PPG-based biometric systems into smartphones, utilizing built-in cameras and flashlights. When the flashlight illuminates the skin tissue, the PPG signals are captured from the resulting reflections with varying intensities by the camera sensors. This method holds significant promise due to its convenience and accessibility. However, the performance of such systems has been hindered by challenges, including low-quality light sources and substantial motion artifacts, leading to low reliability and accuracy.

\begin{figure*}[!t]
\centering
\includegraphics[width=1\textwidth]{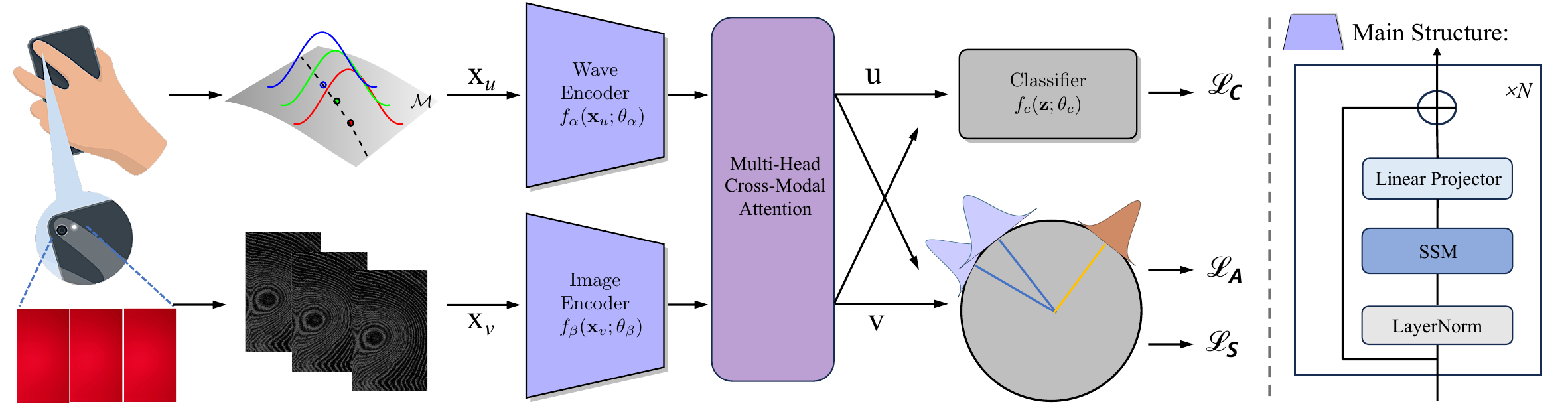}
\caption{{\bf{(Left.)}} The proposed multimodal biometric system. Users place their index finger on the main camera lens with the flashlight on and start recording. After preprocessing the recording videos, the extracted PPG beat waveform and fingerprint image are sequenced and linearly mapped to embeddings $\mathbf{x}_{u}$ and $\mathbf{x}_{v}$. These embeddings are then fed into encoders $f_{\alpha}(\mathbf{x}_{u};\theta_{\alpha})$ and $f_{\beta}(\mathbf{x}_{u};\theta_{\beta})$, and further processed by multi-head cross-modal attention module to obtain latent representations $\mathbf{u}$ and $\mathbf{v}$. Finally, $\mathbf{u}$ and $\mathbf{v}$ are aligned and fused into $\mathbf{z}$, which is passed to the classifier $f_c(\mathbf{z};\theta_c)$ to make the final decision. {\bf{(Right.)}} The main structure of homogeneous encoders. Each encoder consists of $N$ stacks of a deep sequence model, with the core sequence-to-sequence transformation being an SSM, complemented by common neural network components such as residual connections, normalizations, projections, and activations, etc.}
\label{fig_1}
\end{figure*}

In this work, we aim to address these challenges and realize a convenient, accurate, real-time biometric system for daily use. As illustrated in Fig. \ref{fig_1}, the proposed system allows users to place their fingers on the lens while recording video, simultaneously extracting PPG signals and fingerprint images without the need for specialized hardware. We introduce a unified neural network architecture that incorporates two structured state-space model (SSM) encoders \cite{20,21} and a cross-modal attention mechanism, treating both PPG waveforms and fingerprint images as sequences. Moreover, we propose a distribution-oriented contrastive loss to align feature distributions within a shared latent space, enhancing the learning of intrinsic and transferable intermodal relationships. Experimental results demonstrate the system's superior performance across various evaluation metrics in both single-session and dual-session authentication scenarios.

\section{Multimodal Biometric System}
\label{sec:format}
In this section, we introduce the details of the proposed
system, including preprocessing pipeline, neural network architecture, and the alignment algorithm.

\subsection{Preprocessing Pipeline}
The preprocessing consists of two main branches: the signal processing pipeline and the image processing pipeline. When the user's finger fully covers the camera lens, it starts video recordings. Let the RGB video frames be $\mathcal{I}_{rgb}\in \mathbb{R}^{F\times H\times W\times 3}$, which is first transformed into grayscale images $\mathcal{I}_{g}\in \mathbb{R}^{F\times H\times W} $. Here, \( F \) denotes the number of frames, while \( H \) and \( W \) represent the height and width of the image, respectively.

\subsubsection{Signal processing pipeline} We employ the proposed work in \cite{19} to extract PPG beats from recorded grayscale images. The extraction pipeline begins by computing the average pixel intensity of each image frame, generating a raw intensity signal corresponding to the PPG waveform. This raw signal is subsequently detrended and filtered using a low-pass filter with a 4 Hz cutoff frequency. Following this, a beat separation and selection algorithm is applied to isolate individual PPG cycles, ensuring a representative set of beats for further processing. Finally, spline interpolation is used to standardize the length of each beat, facilitating uniform input for subsequent models. The whole process is abstracted as $\mathcal{I}_{g}\in \mathbb{R}^{F\times H\times W}\to \widetilde{U} = \{\widetilde{B}_{1}, \widetilde{B}_{2}, \ldots\}$, where every $\widetilde{B}$ represents a PPG beat waveform.


\subsubsection{Image processing pipeline}
The image processing pipeline is designed to extract the user's fingerprint from the video sequence through the following process. When the flashlight illuminates the finger, the texture of the fingertip creates ridges and contours in the images captured by the camera. Our proposed pipeline begins by applying the CLAHE method \cite{22} to enhance image contrast, followed by the Canny edge detector \cite{23} to extract edges that represent the fingerprint. Finally, fingerprint frames synchronized with the PPG beats are selected, averaged, and downsampled for further analysis. The whole process is abstracted as $\mathcal{I}_{g}\in \mathbb{R}^{F\times H\times W}\to \mathcal{I}_{c}^{\widetilde{U}}=\{ \mathcal{I}_{c}^{\widetilde{B}_{1}},\mathcal{I}_{c}^{\widetilde{B}_{2} },\ldots\}$ with each $\mathcal{I}_{c}^{\widetilde{B}} \in \mathbb{R}^{H_{ds}\times W_{ds}}$, where $H_{ds}$ and $W_{ds}$ are downsampled image height and width, respectively.

For a pair of PPG beat waveform and fingerprint image \(\left(\widetilde{B}, \mathcal{I}_{c}^{\widetilde{B}} \right)\), they are mapped to embeddings \(\mathbf{x}_{u}\) and \(\mathbf{x}_{v}\), respectively, for subsequent processing. Note that \(\mathcal{I}_{c}^{\widetilde{B}}\) should be first converted into a sequence of pixels, where a flatten operation is preferred.

\subsection{Model Architecture}
The proposed architecture as seen in Fig. \ref{fig_1} consist of two parameterized encoders $f_{\alpha}(\mathbf{x}_{u};\theta_{\alpha})$ and $f_{\beta}(\mathbf{x}_{u};\theta_{\beta})$ designed to extract features from $\mathbf{x}_{u}$ and $\mathbf{x}_{v}$. Then a multi-head cross-modal attention module is introduced to learn nuanced latent representations $\mathbf{u}$ and $\mathbf{v}$, and at the end, we have a classifier $f_c(\mathbf{z};\theta_c)$ to make decisions from the generated fused representation $\mathbf{z}$. Without loss of generality, we assume that the embeddings and all latent representations for neural networks share the same dimension, denoted as $d$, for the simplicity of explanation. For instance, we denote the input embeddings as $\mathbf{x}_{u}\in \mathbb{R}^{L_{u}\times d}$ and $\mathbf{x}_{v}\in \mathbb{R}^{L_{v}\times d}$, where $L_{u}$ and $L_{v}$ are sequence lengths. The details of each model/block is explained below:

\subsubsection{Homogeneous SSM encoders}
Given $f_{\alpha}(\mathbf{x}_{u};\theta_{\alpha}):\mathbb{R}^{L_{u}\times d} \to  \mathbb{R}^{L_{u}\times d} $ and $f_{\beta}(\mathbf{x}_{u};\theta_{\beta}): \mathbb{R}^{L_{v}\times d} \to  \mathbb{R}^{L_{v}\times d}$, they consist of $N$ stacks of a deep sequence model that preserves feature dimensions, as shown in the right part of Fig. \ref{fig_1}. Omitting the complementary neural components, the core SSM module \cite{20,21} comprises $d$ parallel 1D sequence-to-sequence transformations. For any one of these parallel transformations, the operation is a discretization of the following continuous system.
\begin{equation}
\begin{aligned}
h'(t) &= \mathbf{A} h(t) + \mathbf{B} x(t) \\
y(t) &= \mathbf{C} h(t) + \mathbf{D} x(t)
\end{aligned}
\label{deqn_ex1}
\end{equation}
where $x(t) \in \mathbb{R}$ is the input, $y(t) \in \mathbb{R}$ is the output, and $h(t)\in \mathbb{R}^{d_{h}}$ is the hidden state vector with dimension of $d_{h}$, containing cumulative information from $x(t)$. Matrices $\mathbf{A}\in\mathbb{C}^{d_{h}\times d_{h}},\mathbf{B}\in\mathbb{C}^{d_{h}},\mathbf{C}\in\mathbb{C}^{d_{h}},\mathbf{D}\in\mathbb{C}^{d_{h}}$ are parameters, respectively. Generally $\mathbf{D} = \mathbf{0}$ for the equivalence of $\mathbf{D} x(t)$ and skip connection, and $\mathbf{A}$ is highly structured for $h(t)$ better capturing long-range dependency. By introducing discretization step parameter $\Delta$ and methods such as zero-order hold (ZOH) to \eqref{deqn_ex1}, we have the sequence-to-sequence transformation
\begin{equation}
\begin{aligned}
h_t &= \overline{\mathbf{A}} h_{t-1} + \overline{\mathbf{B}} x_t\\
y_t &= \mathbf{C} h_t
\end{aligned}
\label{deqn_ex2}
\end{equation}
where $\overline{\mathbf{A}} = \exp(\Delta \mathbf{A})$ and $\overline{\mathbf{B}} = (\Delta \mathbf{A})^{-1} (\exp(\Delta \mathbf{A}) - \mathbf{I}) \cdot \Delta \mathbf{B}$. From \eqref{deqn_ex1} and \eqref{deqn_ex2}, SSM's trainable parameters are $(\Delta, \mathbf{A},\mathbf{B},\mathbf{C})$, which will be updated by gradient descent. 

\subsubsection{Multi-head cross-modal attention} The multi-head cross-modal attention module is designed to capture intricate relationships between different modalities. It accomplishes this by utilizing multiple attention heads, each focusing on distinct aspects of the modal interactions. Let  $\mathbf{g}_{u} = f_{\alpha}(\mathbf{x}_{u};\theta_{\alpha})$ and $\mathbf{g}_{v} = f_{\beta}(\mathbf{x}_{v};\theta_{\beta})$. The attention mechanism \cite{24} is defined as:
\begin{equation}
Attention(\mathbf{Q}, \mathbf{K}, \mathbf{V}) = softmax\left(\frac{\mathbf{Q}\mathbf{K}^\top}{\sqrt{d}}\right)\mathbf{V}
\label{deqn_ex3}
\end{equation}
where $\mathbf{Q}$, $\mathbf{K}$, and $\mathbf{V}$, are the query, key, and value matrices, respectively. For cross-modal attention, we compute attention weights between $\mathbf{g}_{u}$ and $\mathbf{g}_{v}$ and vice versa. The multi-head variant, exemplified by $\mathbf{g}_{u}$ with $\mathbf{g}_{v}$ is
\begin{equation}
\begin{aligned}
MultiHead(\mathbf{g}_{u}, \mathbf{g}_{v}) &= Concat(head_1, \ldots, head_h) \mathbf{W}_O
\end{aligned}
\label{deqn_ex4}
\end{equation}
where each $head_i = Attention(\mathbf{g}_{u} \mathbf{W}_i^Q, \mathbf{g}_{v}\mathbf{W}_i^K, \mathbf{g}_{v}\mathbf{W}_i^V)$, $\mathbf{W}_i^Q$, $\mathbf{W}_i^K$, $\mathbf{W}_i^V$ and $\mathbf{W}_O$ are learnable weight matrices. $Concat$ is concatenation operator. The multi-head cross-modal attention for $\mathbf{g}_{v}$ with $\mathbf{g}_{u}$ is obtained similarly by exchanging $\mathbf{g}_{u}$ and $\mathbf{g}_{v}$ in the expression \eqref{deqn_ex4}, with the other sets of learnable weight matrices. Note that $MultiHead$ also preserves dimensions from input to output, which is consistent with previous illustrations.

After cross-modal attention mechanism, the latent representations $\mathbf{u}$ and $\mathbf{v}$ are $\mathbf{u} = Proj \circ  Pool\circ MultiHead(\mathbf{g}_{u}, \mathbf{g}_{v}): \mathbb{R}^{L_{u}\times d} \to  \mathbb{R}^{d}$ , $\mathbf{v} = Proj \circ  Pool\circ MultiHead(\mathbf{g}_{v}, \mathbf{g}_{u}): \mathbb{R}^{L_{v}\times d} \to  \mathbb{R}^{d}$, respectively. $Pool$ can be any pooling operation over the sequence lengths to aggregate information into a single vector, and $Proj$ is a projection function, which can be linear or nonlinear, that maps the pooled representations to the final latent space.

\subsubsection{Classifier} The classifier $f_c(\mathbf{z};\theta_c)$ is designed to make decisions based on the fused representation $\mathbf{z}$.  $\mathbf{z}$ is derived by combining $\mathbf{u}$ and $\mathbf{v}$ through operations such as concatenation or linear combination. Here we choose fusion operation as an element-wise addition as $\mathbf{u}$ and $\mathbf{v}$ will be normalized and aligned. For authentication, $f_c(\mathbf{z};\theta_c)$ computes binary cross-entropy (BCE) loss $\mathscr{L}_{C}$.

\subsection{Distribution Alignment}
Aligning latent representations from multiple modalities is essential for model generalization. Our work insights multiple biometric modalities as diverse expressions of user identities. Instead of aligning individual instances, we focus on aligning distributions. As shown in Fig. \ref{fig_1}, our method aligns the distributions of two modalities (represented by blue Gaussians) for the target user, while keeping the distributions for other users distinct. This approach recognizes that one PPG waveform may correspond to multiple fingerprint images, and vice versa. 

Given $k$-th batch of \( M \) data points with binary labels sampled from the dataset $\mathcal{D}$ as \(\{(\mathbf{x}_u^{j}, \mathbf{x}_v^{j}), C\}_{j=1:M}^{k}\), where \(\mathbf{x}_u^{j}\) and \(\mathbf{x}_v^{j}\) are the \( j \)-th PPG embedding and fingerprint embedding, respectively, and \( C \in \{C^+, C^-\} \) is the label indicating positive \( C^+ \) and negative \( C^-\) pairs. The model maps \(\{(\mathbf{x}_u^{j}, \mathbf{x}_v^{j}), C\}_{j=1:M}^{k}\) into a unified latent space as \(\{(\mathbf{u}^{j}, \mathbf{v}^{j}), C\}_{j=1:M}^{k}\), where $\mathbf{u}^{j}$ and $\mathbf{v}^{j}$ are corresponding latent representations. Our method works as follows:
\begin{algorithm}[!t]
\caption{}\label{alg:alg1}
\begin{algorithmic}
\STATE 
\STATE {\textbf{Input}:} Dataset of triples \(\{(\mathbf{x}_u, \mathbf{x}_v), C\}\in \mathcal{D}\) and initialization $\mathbf{\Phi}^{0} = \{\theta_{\alpha},\theta_{\beta},\theta_{c}, (\mathbf{W}_i^Q, \mathbf{W}_i^K, \mathbf{W}_i^V, \mathbf{W}_O)_{u/v}, ...\}$, including parameters from projection and fusion layers.
\STATE {\textbf{Output}:} The convergent model parameters $\mathbf{\Phi}$.
\STATE \hspace{0.5cm} \textbf{for} epoch $e=0$ to $E$ \textbf{do}
\STATE \hspace{1cm} \textbf{for} batch $k=1$ to $K$ \textbf{do}
\STATE \hspace{1.5cm}\textbf{Parallel forward} \(\{(\mathbf{x}_u^{j}, \mathbf{x}_v^{j}), C\}_{j=1:M}^{k}\)
\STATE \hspace{1.5cm}$\mathbf{g}_{u}^{j}, \mathbf{g}_{v}^{j}  \gets f_{\alpha}(\mathbf{x}_{u}^{j};\theta_{\alpha}),f_{\beta}(\mathbf{x}_{v}^{j};\theta_{\beta})$
\STATE \hspace{1.5cm}$\mathbf{u}^{j} \gets Proj \circ  Pool\circ MultiHead(\mathbf{g}_{u}^{j}, \mathbf{g}_{v}^{j}) $
\STATE \hspace{1.5cm}$ \mathbf{v}^{j} \gets Proj \circ  Pool\circ MultiHead(\mathbf{g}_{v}^{j}, \mathbf{g}_{u}^{j}) $
\STATE \hspace{1.5cm}\textbf{if} $e==0$ \textbf{do}
\STATE \hspace{2cm}$ \hat{\mu}_{\mathbf{u}}^{k}, \hat{\mu}_{\mathbf{v}}^{k} \gets $ 
\eqref{deqn_ex7} by $\hat{\mu}_{\mathbf{u}}^{0}=0, \hat{\mu}_{\mathbf{v}}^{0}=0$ 
\STATE \hspace{1.5cm}\textbf{else do}
\STATE \hspace{2cm}$ \hat{\mu}_{\mathbf{u}}^{k}, \hat{\mu}_{\mathbf{v}}^{k} \gets $ 
\eqref{deqn_ex7} by $\hat{\mu}_{\mathbf{u}}^{0},\hat{\mu}_{\mathbf{v}}^{0} =\hat{\mu}_{\mathbf{u}}^{K},\hat{\mu}_{\mathbf{v}}^{K}$ at $e-1$
\STATE \hspace{1.5cm}$ \mathscr{L}_{C} \gets f_c(\mathbf{z}(\mathbf{u}^{j},\mathbf{v}^{j});\theta_c)_{j=1:M}$
\STATE \hspace{1.5cm}$ \mathscr{L}_{A},\mathscr{L}_{S} \gets$ by \eqref{deqn_ex5},\eqref{deqn_ex6}
\STATE \hspace{1.5cm}\textbf{Backward} $\nabla_{\mathbf{\Phi}} \mathscr{L}$ by \eqref{deqn_ex8}
\STATE \hspace{1.5cm}$\mathbf{\Phi} \gets Gradient Descent (\nabla_{\mathbf{\Phi}} \mathscr{L})$
\STATE \hspace{1.0cm}\textbf{end for}
\STATE \hspace{0.5cm}\textbf{end for} 
\end{algorithmic}
\label{alg1}
\end{algorithm}      
\subsubsection{Aligning estimates} In the \(k\)-th batch, the representations \(\mathbf{u}^{j}\) and \(\mathbf{v}^{j}\) belonging to the positive label \(C^+\) need to be aligned. Since explicit measures for these representations are unavailable, their first-order moments can be used as alignment indicators. Suppose \(\hat{\mu}_{\mathbf{u}}^{k}\) and \(\hat{\mu}_{\mathbf{v}}^{k}\) be the first-order moment estimates and are known. We can align these moments using the InfoNCE loss\cite{25} as 
        \begin{equation}
        \begin{aligned}
        \mathscr{L}_{A} & =  \log \frac{\exp(sim(\hat{\mu}_{\mathbf{u}}^{k}, \hat{\mu}_{\mathbf{v}}^{k}) / \tau)}{\sum_{\mathbf{v}^{j} \in \hat{\mu}_{\mathbf{v}}^{k} \cup C^{-}} \exp(sim(\hat{\mu}_{\mathbf{u}}^{k},\mathbf{v}^{j}  ) / \tau)} \\
        &+ \log \frac{\exp(sim(\hat{\mu}_{\mathbf{u}}^{k}, \hat{\mu}_{\mathbf{v}}^{k}) / \tau)}{\sum_{\mathbf{u}^{j}  \in \hat{\mu}_{\mathbf{u}}^{k} \cup C^{-}} \exp(sim(\mathbf{u}^{j} , \hat{\mu}_{\mathbf{v}}^{k}) / \tau)} 
        \end{aligned}
        \label{deqn_ex5}
        \end{equation}
         where $sim(\cdot,\cdot)$ is for capturing cosine similarity or spherical geodesic distance between two vectors, and $\tau$ is the temperature parameter that scales the similarity scores. This alignment makes the estimates from positive labels close while pushing any other negative instances apart.
\subsubsection{Controlling spread} For aligning distributions, the first-order moments alone may not be sufficiently representative, as the spread of the distributions is also vital and should be controlled. Therefore, we constrain the interclass variance within the positive pairs as    
        \begin{equation}
        \begin{aligned}
        \mathscr{L}_{S} & = \sum_{(\mathbf{u}^{j},\mathbf{v}^{j}) \in{C^{+}}} \left( \frac{\Vert \mathbf{u}^{j}-\hat{\mu}_{\mathbf{u}}^{k} \Vert}{\Vert \hat{\mu}_{\mathbf{u}}^{k}\Vert} + \frac{\Vert \mathbf{v}^{j}-\hat{\mu}_{\mathbf{v}}^{k} \Vert}{\Vert \hat{\mu}_{\mathbf{v}}^{k} \Vert} \right)
        \end{aligned}
        \label{deqn_ex6}
        \end{equation}
        where $\Vert \cdot \Vert$ represents the squared L$_{2}$-norm as $\Vert \cdot \Vert_2^2$. Note this loss captures the normalized variances. By combining \eqref{deqn_ex5}\eqref{deqn_ex6}, the losses steer the formation of two aligned clusters in the unified latent space.

\subsubsection{EMA updating} The derivation of the estimates above in \eqref{deqn_ex5} is cored by a cumulative updating rule across batches. We utilize exponential moving averages (EMA) to achieve this by weighting successive batch means. This ensures that the estimates evolve to more accurately represent the central tendency of positive samples as more data is processed. Moreover, EMA aids in capturing a more global perspective, thereby reducing the risk of overfitting to the peculiarities of individual batches \cite{26}. This process is presented recursively as
\begin{equation}
    \begin{aligned}
    (\hat{\mu}_{\mathbf{u}}^{k+1}, \hat{\mu}_{\mathbf{v}}^{k+1}) = & (\alpha, \beta) \odot ( \hat{\mu}_{\mathbf{v}}^{k},  \hat{\mu}_{\mathbf{v}}^{k}) \\
    & + (1 - \alpha, 1 - \beta) \odot (\mathbf{\overline{u}}_{C^{+}}^{k+1}, \mathbf{\overline{v}}_{C^{+}}^{k+1})
    \end{aligned}
    \label{deqn_ex7}
    \end{equation}
    where $(\mathbf{\overline{u}}_{C^{+}}^{k+1}, \mathbf{\overline{v}}_{C^{+}}^{k+1}) = \frac{1}{N_{C^{+}}} \sum_{(\mathbf{u}^{j},\mathbf{v}^{j})_{\in {C^{+}}}} (\mathbf{u}^{j},\mathbf{v}^{j})_{k+1}$ represent the average of positive samples in $k+1$-th batch, $N_{C^{+}}$ is the number of positive samples in this batch. Here $\alpha$ and $\beta$ are decaying factors. $\odot$ is Hadamard product. The final loss is the combination as 
    \begin{equation}
    \begin{aligned}
    \mathscr{L} = \mathscr{L}_{C} + \lambda_A \mathscr{L}_{A} + \lambda_S \mathscr{L}_{S}
    \end{aligned}
    \label{deqn_ex8}
    \end{equation}
    where $\lambda_A$ and $\lambda_S$ are penalty factors. The training algorithm for the model is outlined in Algorithm \ref{alg:alg1}.

\section{Experiments}
\label{sec:typestyle}
This section presents the experimental results demonstrating the effectiveness of the proposed method. First, we introduce new datasets consisting of fingertip videos collected using smartphone cameras. Second, authentication and ablation study results are reported using accuracy (ACC) and equal error rate (EER) as evaluation metrics. 

\subsection{Dataset Collection}
We collected two datasets: a single-session dataset, \textit{V1}, consisting of 40 subjects (30 males, 10 females), and a two-session dataset, \textit{V2}, comprising 27 subjects (16 males, 11 females) with an average interval of 7 days between sessions. The subjects, aged 24-40 and with no cardiovascular diseases or fingerprint damage, were recorded using a Vivo Y55 5G smartphone at 1080p resolution and 60 fps, with the flashlight enabled. The dataset totals over 90 GB. During each session, subjects were instructed to remain still while completing nine 30-second recordings, with breaks in between to introduce variation.

\begin{table}[!t]
    \caption{The Average ACC and EER for Single Session Evaluation}
    \label{tab:table1}
    \centering
    \begin{tabular} {l c c c c c} 
        \toprule
        Method & Type & Subjects & Model & ACC $\uparrow$ & EER $\downarrow$ \\
        \midrule
        Kari.\cite{27}  & IR    & 37 & SVM  & $99.75\%$    & $1.46\%$  \\
        Hwang \cite{10} & IR    & 37 & CNN-LSTM & $100\%$     & $0.1\%$  \\
        See red \cite{19} & Video & 14& SVM & $86.6\%$ & $13.9\%$ \\
        \textbf{Ours(\textit{V1})} & Video & 40 & - & $\mathbf{100\%}$ & $\mathbf{0.1\%}$ \\
        \bottomrule
    \end{tabular}
\end{table}

\begin{table}[!t]
\caption{The Average ACC and EER for Two Session Evaluation\label{tab:table2}}
\centering
    \begin{tabular} {l c c c c c c} 
        \toprule
        Method & Type & Dataset & Subjects & ACC $\uparrow$ & EER $\downarrow$\\
        \midrule
        San.\cite{28} & IR   & Berry & 24   & -     & $19.1\%$  \\
        San.\cite{28}  & IR   & Nomin & 24   & -     & $23.2\%$  \\
        Hwang\cite{10}  & IR   & Biosec1  & 31  & $87\%$    & $13\%$  \\
        Hwang\cite{10} & IR   & Biosec2  & 100  & $87.1\%$    & $12.9\%$  \\
        See red \cite{19} & Video & -  & 14  & $85.3\%$ & $21.7\%$ \\
        \textbf{Ours} & Video & \textit{V2}  & 27 & $\mathbf{94.6\%}$ & $\mathbf{6.9\%}$\\
        \bottomrule
    \end{tabular}
\end{table}

\subsection{Authentication Performance}
Table 1 and Table 2 demonstrate the authentication performance of the proposed system. We let $\widetilde{B}\in \mathbb{R}^{300 \times 1}$ and fingerprint image $\mathcal{I}_{c}^{\widetilde{B}}\in \mathbb{R}^{64 \times 64} $, respectively. The hyperparameter of $300$ specifies the beat waveform length, while $64\times64$ represents the image resolution. These values are chosen to balance complexity and performance, especially for the images, where flattening leads to a quadratic increase in sequence length.
Moreover, the data augmentation method from \cite{10} is applied to the PPG beat waveforms, while conventional image transformation methods, such as flipping, rotation, affine transformation, cropping, and adding noise, are applied to the fingerprint images. Furthermore, we duplicate the data samples of the true user by allowing PPG beat waveforms and fingerprint images to be randomly matched in a many-to-many relationship instead of a one-to-one relationship, forming multiple pairs.

Algorithm \ref{alg:alg1} is applied to the embeddings $\mathbf{x}_{u} \in \mathbb{R}^{300 \times 128}$ and $\mathbf{x}_{v} \in \mathbb{R}^{4096 \times 128}$, which are linearly mapped from \(\left(\widetilde{B}, \mathcal{I}_{c}^{\widetilde{B}} \right)\). The encoders $f_{\alpha}(\mathbf{x}_{u};\theta_{\alpha})$ and $f_{\beta}(\mathbf{x}_{v};\theta_{\beta})$ are S4D \cite{21}  variants with a hidden state dimension of $d_{h} = 64$ for SSM and $N=2$ stacks of a deep sequence model. The number of heads for the cross-modal attention module $MultiHead$ is set to 4 for both $\mathbf{u}$ and $\mathbf{v}$. $Pool$ is average pooling, and $Proj$ is a linear projection function. The fusion function $\mathbf{z}$ is chosen as element-wise addition, and $f_c(\mathbf{z};\theta_c)$ computes a weighted BCE loss to handle the imbalanced classes.
The number of epochs is set to 200, with a batch size of 256, and Adam is used as the optimizer. The decay factors $\alpha$ and $\beta$ for equation \eqref{deqn_ex7}, and penalty factors $\lambda_{A}$ and $\lambda_{C}$ for equation \eqref{deqn_ex8} are empirically determined as $0.9, 0.9, 0.8$, and $0.05$, respectively. The overall Algorithm \ref{alg:alg1} is implemented using PyTorch and runs on Nvidia RTX 4090D GPUs.

As shown in Table 1, our method significantly improves authentication performance in the single-session scenario. We benchmarked against the See Red method \cite{19}, the only camera-based PPG approach, and for a comprehensive comparison, we also evaluated PPG authentication systems using IR devices, as referenced in \cite{10,27}, on the PRRB dataset, which is comparable in size to our\( V1 \)dataset. Table 1 demonstrates that our system achieves state-of-the-art (SOTA) results, with an ACC of \( 100\% \) and an EER of \( 0.1\% \), marking a significant improvement over the camera-based See Red method’s re-evaluated performance, which achieved an ACC of \( 86.6\% \) and an EER of \( 13.9\% \), and is comparable to the best results obtained with IR-based PPG systems. Additionally, Table 2 further highlights the robustness of our method in two-session scenarios, demonstrating that our approach maintains superior performance against \cite{19}, which uses the same modality, as well as other IR-based PPG models. This confirms the effectiveness and generalizability of the proposed approach for the authentication task. 



\subsection{Ablation Study}
To evaluate the effectiveness of our proposed model, we conducted an ablation study using four SOTA baseline models:
\begin{enumerate}
    \item \textit{SSM fingerprint/PPG Baseline:} An SSM-based model that utilizes only PPG signals or fingerprints.
    \item \textit{CNN-LSTM PPG Baseline:} A model using the CNN-LSTM architecture from \cite{10}, trained solely on PPG signals.
    \item \textit{ResNet-18 Fingerprint Baseline:} A model leveraging the ResNet-18 architecture, trained exclusively on fingerprint data.
\end{enumerate}

Each model was trained for a maximum of 80 epochs with a batch size of 32, allowing us to systematically assess the contribution of individual modalities (PPG vs. fingerprint) and compare different architectural frameworks (SSM vs. CNN-LSTM vs. ResNet-18) to the overall system performance.
As shown in Table \ref{tab:table3}, combining the SSM model with the camera-based PPG delivers an acceptable EER of 16.7\%. However, integrating fingerprint data enhances authentication performance, yielding substantially lower error rates. Additionally, the high EER observed with the CNN-LSTM model suggests that the inherent characteristics of camera-based PPG signals are easier to discover using the SSM model, but more challenging to capture than those obtained from IR-based sensors. This highlights the increased complexity of extracting robust features from camera-based PPG signals, emphasizing the need for more sophisticated processing techniques to improve accuracy. 



\section{Conclusion}
\label{sec:majhead}
\begin{table}[!t]
\caption{Ablation study\label{tab:table3}}
\centering
    \begin{tabular} {l c c c c } 
        \toprule
        Method & Modality & EER(V1) & EER(V2) & FLOPs\\
        \midrule
        SSM\cite{21} & PPG   & $16.7\%$  & $24\%$ & $17.14$M  \\
        SSM\cite{21}  & Fingerprints   & $1.31\%$ &  $10.1\%$ & $20.05$M    \\
        CNN-LSTM\cite{10}  & PPG   & $45.7\%$  & -  & -   \\
        ResNet-18\cite{29} & Fingerprints   &  $1.65\%$  &  $11\%$& 1.8G \\
        \bottomrule
    \end{tabular}
\end{table}
In this paper, we present a multimodal biometric system that combines camera-based PPG signals with fingerprint data to enhance human verification accuracy. Using a neural network with state-space model encoders and a cross-modal attention mechanism, the system effectively processes and aligns these biometric features within a unified latent space. Experimental results demonstrate strong performance in both single-session and two-session scenarios, underscoring its potential for practical applications in secure authentication.

%

\vfill\pagebreak


\end{document}